\documentclass[twoside,11pt]{article}

\usepackage{jmlr2e}
\usepackage{amsmath, bm}
\usepackage{url}
\usepackage{booktabs}
\usepackage{listings}
\usepackage{xcolor}
\usepackage{subcaption}

\definecolor{codegreen}{rgb}{0,0.6,0}
\definecolor{codegray}{rgb}{0.5,0.5,0.5}
\definecolor{codepurple}{rgb}{0.58,0,0.82}
\definecolor{backcolour}{rgb}{0.95,0.95,0.92}

\lstdefinestyle{mystyle}{
    backgroundcolor=\color{backcolour},   
    commentstyle=\color{codegreen},
    keywordstyle=\color{magenta},
    numberstyle=\tiny\color{codegray},
    stringstyle=\color{codepurple},
    basicstyle=\ttfamily\footnotesize,
    breakatwhitespace=false,         
    breaklines=true,                 
    captionpos=b,                    
    keepspaces=true,                 
    numbers=left,                    
    numbersep=5pt,                  
    showspaces=false,                
    showstringspaces=false,
    showtabs=false,                  
    tabsize=2
}

\jmlrheading{Under Review}{2022}{1-11}{10/22}{AB/CD}{Thanh Tung Khuat and Bogdan Gabrys}

\ShortHeadings{hyperbox-brain: A Toolbox for Hyperbox-based Machine Learning Algorithms}{Khuat and Gabrys}
\firstpageno{1}

\begin{document}

\title{\texttt{hyperbox-brain}: A Python Toolbox for Hyperbox-based Machine Learning Algorithms}

\author{\name Thanh Tung Khuat \email thanhtung.khuat@uts.edu.au \\
       \addr Complex Adaptive Systems Lab, The Data Science Institute\\
       University of Technology Sydney\\
       Sydney, NSW 2007, Australia
       \AND
       \name Bogdan Gabrys \email bogdan.gabrys@uts.edu.au \\
       \addr Complex Adaptive Systems Lab, The Data Science Institute\\
       University of Technology Sydney\\
       Sydney, NSW 2007, Australia}

\editor{ABC}

\maketitle

\begin{abstract}
Hyperbox-based machine learning algorithms are an important and popular branch of machine learning in the construction of classifiers using fuzzy sets and logic theory and neural network architectures. This type of learning is characterised by many strong points of modern predictors such as a high scalability, explainability, online adaptation, effective learning from a small amount of data, native ability to deal with missing data and accommodating new classes. Nevertheless, there is no comprehensive existing package for hyperbox-based machine learning which can serve as a benchmark for research and allow non-expert users to apply these algorithms easily. \texttt{hyperbox-brain} is an open-source Python library implementing the leading hyperbox-based machine learning algorithms. This library exposes a unified API which closely follows and is compatible with the renowned \texttt{scikit-learn} and \texttt{numpy} toolboxes. The library may be installed from Python Package Index (PyPI) and the \texttt{conda} package
manager and is distributed under the GPL-3 license. The source code, documentation, detailed tutorials, and the full descriptions of the API are available at \url{https://uts-caslab.github.io/hyperbox-brain}.
\end{abstract}


\section{Introduction}
This \texttt{hyperbox-brain} toolbox has been developed by the researchers within the Complex Adaptive Systems laboratory at the University Technology Sydney. It is a result of many years of developing versatile machine learning algorithms with hyperboxes as the foundational representation element at their core.

Hyperbox-based machine learning algorithms use min-max hyperboxes as their fundamental building blocks to partition the sample space into various regions. A collection of hyperboxes representing the same class can form the regions of arbitrary shape and complexity. Each min-max hyperbox is usually characterised by the minimum and maximum vertices together with a fuzzy membership function acting as a distance or similarity measure. During the training procedure, these hyperboxes are formed, as needed, and adjusted to accommodate the incoming input samples based on the degree-of-fit of each input pattern to given hyperboxes expressed by their membership values. The use of hyperboxes for learning systems can deal effectively with the pattern classification and clustering problems \citep{khga20}. Moreover, this kind of learning exposes numerous essential properties for lifelong machine learning systems \citep{ha01, crca20} such as scalability, explainability, incremental adaptation in dynamically changing environments, continuously learning ability from a limited amount of data and new samples, absorption of new knowledge as well as accommodating new classes with a better ability to avoid catastrophic forgetting, due to their local data cluster representations, and capability to manage the stability-plasticity dilemma \citep{mcco89}.

\begin{figure}[!ht]
    \centering
    \includegraphics[width=0.55\textwidth]{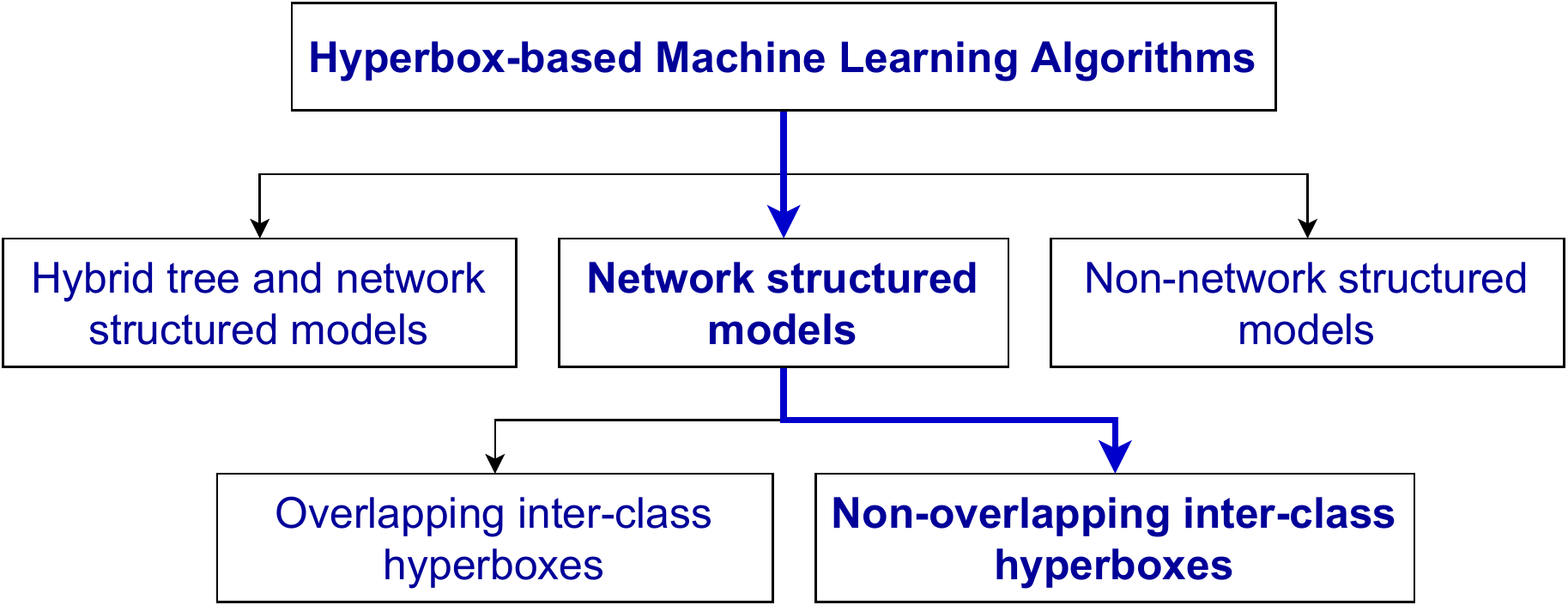}
    \caption{Taxonomy of hyperbox-based machine learning algorithms.}
    \label{fig_1}
\end{figure}

According to a recent survey \citep{khru21}, hyperbox-based machine learning algorithms can be divided into three main groups as illustrated in Fig. \ref{fig_1}. The first group includes network structured learning models. This group contains two types of learning algorithms. The first sub-group allows the occurrence of overlapped areas among hyperboxes representing different class labels, while the hyperboxes belonging to different classes generated by the algorithms in the other sub-group are not allowed to overlap with each other. The second main group consists of hybrid tree and network structured models. The last group includes non-network structured models. The learning algorithms supported by this initial release of the \texttt{hypebox-brain} library primarily come from the network structured models with non-overlapping inter-class hyperboxes. 

\vspace{-0.4cm}

The use of hyperbox fuzzy sets as fundamental representation blocks for machine learning based classifiers dates back to the early 1990s with the most prominent early works including a fuzzy min-max neural network (FMNN) \citep{simpson1992fuzzy} proposed by Simpson and the variations and extension to the Adaptive Resonance Theory (ART) such as Fuzzy ART \citep{Carpenter1991fuzzy} and ARTMAP \citep{Carpenter1991artmap} proposed by Carpenter et al.

This \texttt{hyperbox-brain} library focuses mainly on the FMNN and its improved, later variants such as the general fuzzy min-max neural network (GFMMNN) \citep{gabrys2000general}. Apart from scalability, explainability, incremental adaptation, and continuous learning from limited input samples, learning algorithms of GFMMNN also exhibit unique learning abilities such as learning from interval input data, native learning ability from a mixed labelled and unlabelled training sets \citep{gabrys2000general, gabrys2002agglomerative}, directly learning from data with missing values without the need for data imputation \citep{Gabrys02missing}, combining all resulting hyperboxes in an ensemble model into a single model \citep{Gabrys04multi, gabrys2002combining}, combination of multiple decision trees into a single interpretable hyperbox-based model \citep{Eastwood11}, and the capability of growing and including new classes of data without retraining the whole classifier \citep{Gabrys99water}. All of these unique learning characteristics are supported by the library, and their details will be presented in the next sections.

Although many hyperbox-based machine learning algorithms have been developed over the years with many very recent examples \citep{khru21}, there is no comprehensive software library gathering them in one convenient package allowing their easy usage, benchmarking and further development. Therefore, this paper presents a \texttt{scikit-learn} compatible hyperbox-based machine learning library in Python to fill this gap and serve as a facilitator for further research and applications in this field.

\section{Overview and Design}
To achieve a high performance when applying machine learning algorithms for real-world problems, it is necessary to combine learning algorithms with a hyperparameter search, cross-validation, and feature engineering techniques at a large scale. The \texttt{hyperbox-brain} library, therefore, is designed to be compatible with the \texttt{scikit-learn} toolbox \citep{pedregosa2011scikit} to take advantage of the availability of cross-validation, feature transformation, hyperparameter optimisation, and model evaluation methods. As a result, each model in \texttt{hyperbox-brain} inherits from \texttt{BaseEstimator} or \texttt{BaseEnsemble} and \texttt{ClassifierMixin} in the module \texttt{sklearn.base}. By this inheritance mechanism, the library can set hyperparameters using \texttt{set\_params()}, train a model using \texttt{fit(X, y)}, make a prediction via \texttt{predict(X)}, and evaluate the trained model on a hold-out dataset using \texttt{score()}. To integrate with \texttt{scikit-learn}, \texttt{hyperbox-brain} employs \texttt{numpy}'s structured arrays \citep{harris2020array} whose data type is a combination of simpler data types.

This library aims to provide users with a wide range of learning algorithm categories suitable for addressing different ML problems  and is therefore organised into the following modules:\\
\textit{base}: Providing base classes and functions for all hyperbox-based models in the library.\\
\textit{mixed data learner}: Containing the specialised estimators which can work on mixed-attribute data. However, categorical features given in a text form can also be encoded by various encoding methods so that they can be processed by the following learning algorithms for the numerical data only \citep{khuat2021depth}.\\
\textit{incremental learner}: Including estimators for numerical data which use the instance based incremental learning approaches.\\
\textit{batch learner}: Comprising hyperbox-based learning algorithms for numerical data using batch learning approaches.\\
\textit{multigranular learner}: Containing classifiers for numerical data using the multigranularity learning methods.\\
\textit{ensemble learner}: Including the combination of hyperbox-based learners integrated with various ensemble learning methods.\\
\textit{utils}: Containing utility functions associated with unit tests which can be executed on all supported Python versions by the continuous integration workflow.

\section{The \texttt{hyperbox-brain} Library at a Glance}
The \texttt{hyperbox-brain} library is intended for researchers and practitioners as an easily accessible toolbox of hyperbox-based machine learning algorithms. This library is implemented in Python, providing numerous learning algorithms using hyperboxes as fundamental building blocks for solving classification and clustering problems. Our purpose is to create an easy-to-use package which may be extended by the community, while also providing essential functionality and characteristics to enable researchers and practitioners to benchmark, reproduce, further develop and apply this type of algorithms for their problems.

\subsection{Supported Hyperbox-based Machine Learning Algorithms}
The \texttt{hyperbox-brain} library currently includes 18 hyperbox-based algorithms, which are used to train network-structured learning models without enabling the overlapped areas among inter-class hyperboxes. All of them are original and improved learning algorithms for the fuzzy min-max neural network \citep{simpson1992fuzzy} and the general fuzzy min-max neural network \citep{gabrys2000general}. Of these 18 algorithms, there are three mixed-data learners and 15 learners for numerical data. Out of 15 numerical data learners, there are six instance-incremental learners, two batch learners, six ensemble learners, and one multi-granularity learner. These algorithms are summarised in Table \ref{table_1}.

\begin{table}[!ht]
    \centering
    \scriptsize{
    \begin{tabular}{p{7cm}ccl}
        \toprule
        \textbf{Model} & \textbf{Feature type} & \textbf{Model type} & \textbf{Learning type}\\
        \midrule
        EIOL-GFMM \citep{khuat2020online} & Mixed & Single & Instance-incremental \\ \midrule
        Freq-Cat-Onln-GFMM \citep{khga20} & Mixed & Single & Batch-incremental \\ \midrule
        OneHot-Onln-GFMM \citep{khuat2020online} & Mixed & Single & Batch-incremental \\ \midrule
        Onln-GFMM \citep{gabrys2000general} & Numerical & Single & Instance-incremental \\ \midrule
        IOL-GFMM \citep{khch20} & Numerical & Single & Instance-incremental \\ \midrule
        FMNN \citep{simpson1992fuzzy} & Numerical & Single & Instance-incremental \\ \midrule
        EFMNN \citep{mohammed2014enhanced} & Numerical & Single & Instance-incremental \\ \midrule
        KNEFMNN \citep{mohammed2017improving} & Numerical & Single & Instance-incremental \\ \midrule
        RFMNN \citep{al2020refined} & Numerical & Single & Instance-incremental \\ \midrule
        AGGLO-SM \citep{gabrys2002agglomerative} & Numerical & Single & Batch \\ \midrule
        AGGLO-2 \citep{gabrys2002agglomerative} & Numerical & Single & Batch \\ \midrule
        MRHGRC \citep{khuat2021effective} & Numerical & Single & Multi-Granularity \\ \midrule
        Decision-level Bagging \citep{gabrys2002combining} & Numerical & Combination & Ensemble \\ \midrule
        Decision-level Bagging + hyperparameter optimisation for base learners & Numerical & Combination & Ensemble \\ \midrule
        Model-level Bagging \citep{gabrys2002combining} & Numerical & Combination & Ensemble \\ \midrule
        Model-level Bagging + hyperparameter optimisation for base learners & Numerical & Combination & Ensemble \\ \midrule
        Random hyperboxes \citep{khuat2021random} & Numerical & Combination & Ensemble \\ \midrule
        Random hyperboxes + hyperparameter optimisation for base learners & Numerical & Combination & Ensemble \\
        \bottomrule
    \end{tabular}
    \caption{Hyperbox-based machine learning algorithms are supported by \texttt{hyperbox-brain}.}
    \label{table_1}
    }
\end{table}

\subsection{Typical Features of the Library}
\textbf{\texttt{scikit-learn} compatibility}: The library is designed to be compatible with and benefits from the \texttt{scikit-learn} toolbox's many features including hyperparameter search, model section and evaluation techniques as well as the pipeline composition approaches (see Section \ref{compatibility}). Moreover, the library can be compatible with other hyperparameter optimisation libraries which may be integrated with \texttt{scikit-learn} such as \texttt{hyperopt} \citep{bergstra2013making}.\\
\textbf{Explainability}: One of the interesting characteristics of the use of hyperbox fuzzy sets for building pattern classifiers is the explainability of the predicted results (see the Appendix A for more details). The library supports this functionality by possible parallel coordinates plots based visualisation of representative hyperboxes from different classes together with an input pattern to be classified.\\
\textbf{Capability of directly handling missing data}: General fuzzy min-max neural networks supported by the library have the ability to handle the classification of inputs with missing data directly without the need for replacing or imputing missing values as in other classifiers \citep{Gabrys02missing}.\\
\textbf{Combination of multiple models at the model level}: Learning algorithms for the GFMMNN in the library can combine multiple decision trees \citep{Eastwood11} or resulting hyperboxes generated by multiple hyperbox-based models \citep{gabrys2002combining} into a single model. This feature contributes to the increase of explainability of ensemble models.\\
\textbf{Data editing and pruning approaches}: By integrating the repeated cross-validation methods provided by \texttt{scikit-learn} and hyperbox-based learning algorithms, evidence from training multiple models can be used for identifying which points from the original data set or the hyperboxes from the generated multiple models should be retained and those that should be edited out \citep{Gabrys01editing} or pruned \citep{Gabrys04multi} before further processing.\\
\textbf{Native ability to learn from both labelled and unlabelled data}: One of the outstanding features of learning algorithms for the GFMMNN is the ability to form classification boundaries between known classes and ability to cluster data and represent them as hyperboxes when labels are not available in the data \citep{gabrys2000general, gabrys2002agglomerative}. Unlabelled hyperboxes may be then labelled based on the evidence of incoming input patterns.\\
\textbf{Ability to learn from new classes in an incremental manner}: Incremental learning algorithms of hyperbox-based models provided in the library can grow and include new classes of data without the need for retraining the whole classifier \citep{Gabrys99water}. Incremental learning algorithms themselves can develop new hyperboxes to represent clusters of new data with potentially new labels both in the middle of normal training process and in the operating time where the initial training has been completed. This characteristic is a key feature for life-long learning systems.\\
\textbf{Documentation and tutorials}: A comprehensive documentation is developed using \texttt{sphinx} and \texttt{numpydoc} and is provided to users via the \textit{Read the Doc} platform\footnote{\url{https://hyperbox-brain.readthedocs.io/en/latest/}}. It provides a detailed API reference, essential background, and a wide range of tutorials and examples\footnote{\url{https://hyperbox-brain.readthedocs.io/en/latest/tutorials/tutorial\_index.html}} under the interactive \textit{Jupyter notebook} to allow new users to explore how the classifiers in the library are used for solving classification problems.\\
\textbf{Build robustness}: The library uses GitHub Actions for continuous integration. Automated scripts are used for automated testing and building the library under different versions of Python and operating systems. Tests are executed for each commit made to master branch or when a pull request is opened.\\
\textbf{Quality assurance}: Code in the project follows the PEP8 style standard for Python. In addition, essential utility functions and code blocks with high complexity are accompanied with a set of unit tests. Furthermore, continuous integration is conducted to guarantee backward compatibility and integrate new code in an easy fashion.\\
\textbf{Community-based development}: We welcome the contributions from the community to the library via collaborative tools such as Git and GitHub. We provide a documented contribution guideline\footnote{\url{https://hyperbox-brain.readthedocs.io/en/latest/developers/contributing.html}} to describe various ways that contributors can join and contribute to the library. In addition, GitHub's issue tracker and discussion are used to discuss ideas and report bugs regarding the library. The \texttt{hyperbox-brain} library is distributed under the GPL-3.0 license.

\subsection{Installation and Usage}\label{usage}
The \texttt{hyperbox-brain} toolbox can be downloaded and installed via \texttt{PyPI} using the command \texttt{pip install hyperbox-brain} or from \texttt{conda-forge} using the command \texttt{conda install -c conda-forge hyperbox-brain}. It is also possible to clone the source code directly from GitHub\footnote{\url{https://github.com/UTS-CASLab/hyperbox-brain}}. In this case, the library can be installed by executing the existing setup script in the root directory through the command \texttt{python setup.py install}. Once installed, all available hyperbox-based algorithms and functions in the library can be accessed via importing the desirable class within the \texttt{hbbrain} module. A simple example of fitting and assessing a model in \texttt{hyperbox-brain} is given below. More elaborate examples and tutorials can be accessed via the online documentation.

\begin{footnotesize}
\begin{lstlisting}[language=Python]
>>> from sklearn.datasets import load_iris
>>> from sklearn.preprocessing import MinMaxScaler
>>> from sklearn.model_selection import train_test_split
>>> from sklearn.metrics import accuracy_score
>>> from hbbrain.numerical_data.incremental_learner.onln_gfmm import OnlineGFMM
>>> # Load dataset
>>> X, y = load_iris(return_X_y=True)
>>> # Normalise features into [0, 1] as required by hyperbox-based models
>>> scaler = MinMaxScaler()
>>> scaler.fit(X)
MinMaxScaler()
>>> XX = scaler.transform(X)
>>> # Split data into training and testing sets
>>> X_train, X_test, y_train, y_test = train_test_split(XX, y, test_size=0.3, random_state=42)
>>> # Training a model
>>> clf = OnlineGFMM(theta=0.1).fit(X_train, y_train)
>>> # Make prediction
>>> y_pred = clf.predict(X_test)
>>> acc = accuracy_score(y_test, y_pred)
>>> print(f'Accuracy = {acc * 100: .2f}%')
Accuracy =  97.78%
\end{lstlisting}
\end{footnotesize}

\section{\texttt{scikit-learn} Compatibility} \label{compatibility}
A critical property of the compatibility of \texttt{hyperbox-brain} with \texttt{scikit-learn} is the inheritance and usage of hyperparameter optimisation, model selection and evaluation, and pipeline functionalities. For example,
\begin{itemize}
    \item \texttt{train\_test\_split} of \texttt{scikit-learn} can be used with hyperbox-based models of \texttt{hyperbox-brain} as shown in the example in subsection \ref{usage}.
    \item \texttt{cross\_val\_score} method of \texttt{scikit-learn} can be transparently applied to hyperbox-based models for cross-validation evaluation as below:
\begin{footnotesize}
\begin{lstlisting}[language=Python]
>>> # Instantiating a hyperbox-based model
>>> clf = OnlineGFMM(theta=0.1)
>>> from sklearn.model_selection import cross_val_score
>>> # usage cross_val_score on the hyperbox-based model
>>> cross_val_score(clf, XX, y, cv=5)
array([0.96666667, 0.96666667, 0.86666667, 0.9, 1.])
\end{lstlisting}
\end{footnotesize}
    \item Hyperparameter search functions such as \texttt{grid\_search} and \texttt{random\_search} can be used directly for models in \texttt{hyperbox-brain} as described in the online examples\footnote{\url{https://hyperbox-brain.readthedocs.io/en/latest/tutorials/hyperparameter\_opt.html}}.
    \item Hyperbox-based estimators can be integrated with \texttt{Pipeline} of \texttt{scitkit-learn} to form a combination of feature engineering methods and classifiers as various examples shown in the online tutorials\footnote{\url{https://hyperbox-brain.readthedocs.io/en/latest/tutorials/pipline_integration.html}}.
\end{itemize}

\section{Conclusion}
\texttt{hypberbox-brain} is a free licensed Python toolbox which implements popular hyperbox-based machine learning models. With the high compatibility with \texttt{scikit-learn} API, this library provides users with an easy-to-use package of algorithms which can be integrated with existing cross-validation, pipeline, model selection and evaluation techniques to formulate powerful data analytics pipelines. As shown in Fig. \ref{fig_1}, there are many existing hyperbox-based learning algorithms in the literature, therefore, we are continuously adding new models, enhancing usability, code quality, unit tests, documents, and tutorials. Finally, we strongly encourage the contributions of the community to expand this free library for the benefit of both researchers and practitioners interested in and able to benefit from this type of ML algorithms.



\appendix

\section*{Appendix A. Explainability of Predicted Outcomes} \label{explanation}
This appendix is dedicated to clarify the explanability of hyperbox-based learning algorithms for the predicted results of a given input pattern through different types of visualisation supported by the \texttt{hyperbox-brain} library.

\begin{figure}[!ht]
     \begin{subfigure}[b]{0.48\textwidth}
     \includegraphics[width=0.92\linewidth]{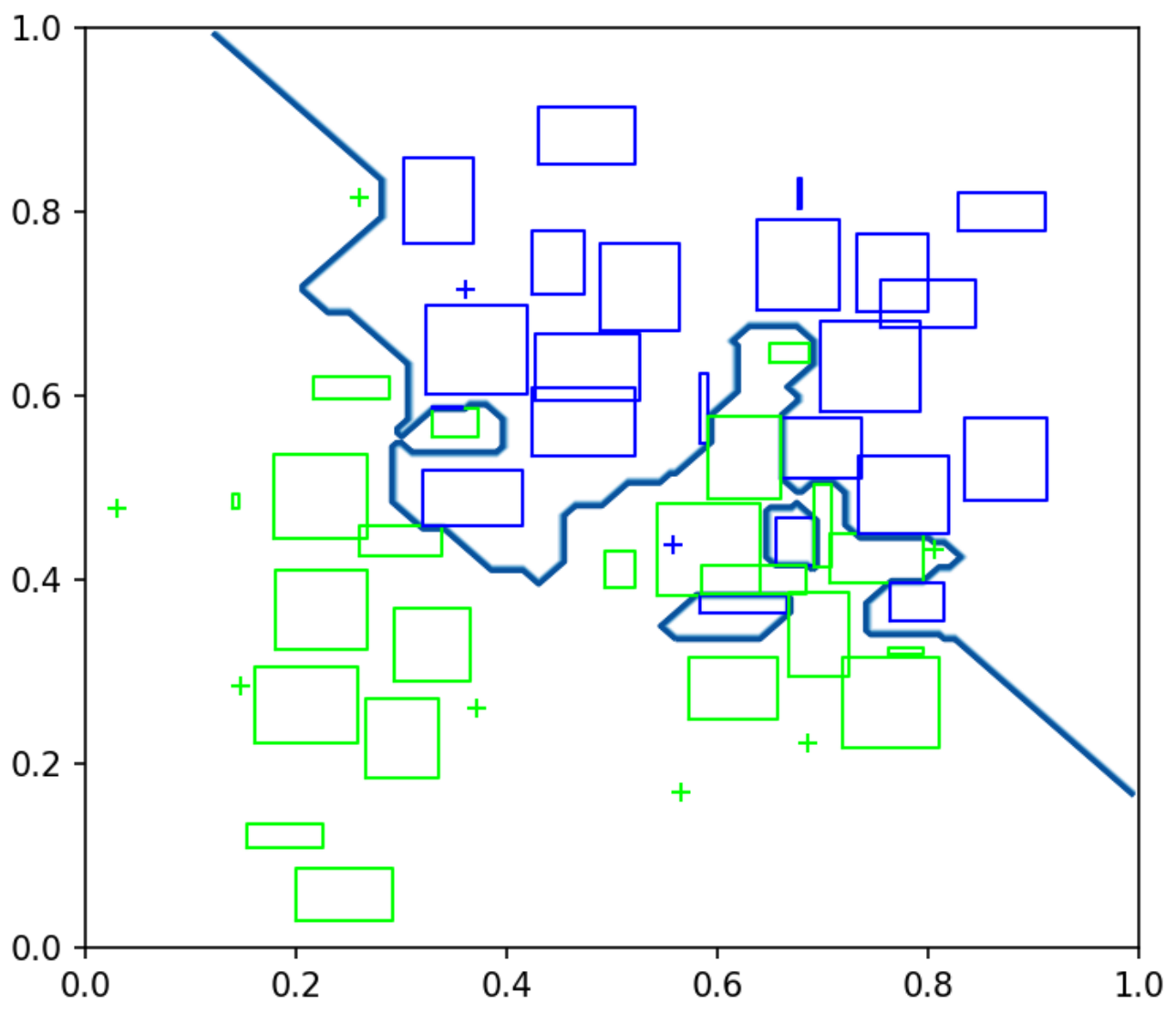}
     \caption{}\label{hyperboxes_boundaries}
     \end{subfigure}
    \begin{subfigure}[b]{0.48\textwidth}
        \includegraphics[width=0.92\linewidth]{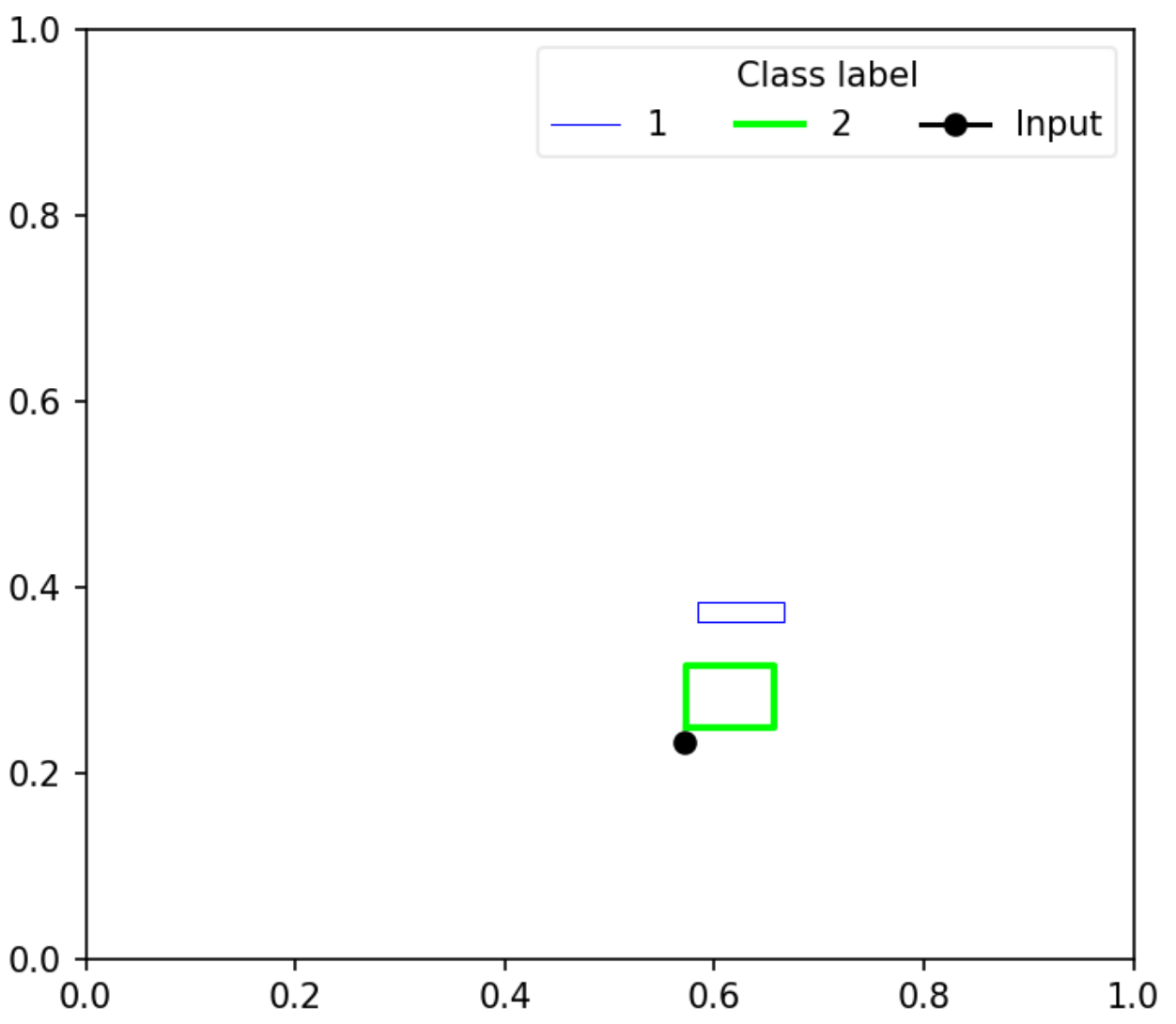}
        \caption{}\label{expanation_2d}
    \end{subfigure}
\caption{(a) Visualisation of two dimensional hyperboxes and the decision boundaries of a trained GFMMNN. (b) Illustration of two dimensional representative (i.e. winning) hyperboxes from two distinct classes used for the classification of an input pattern as a part of possible suggested decision explanation approach.}
\label{errordifsamplesyn}
\end{figure}

For two dimensional training samples, the library provides a functionality to show the generated hyperboxes and decision boundaries among classes of a trained model, e.g., GFMMNN, by a hyperbox-based machine learning algorithm as in Fig. \ref{hyperboxes_boundaries}. For a given two dimensional input pattern and a trained hyperbox-based model, the library shows representative hyperboxes of all class labels joining the prediction to make the predicted class in a two dimensional plane as shown in Fig. \ref{expanation_2d}. The predicted class for an unseen pattern is the same with the hyperbox which has the highest membership value to that input pattern. Let's take the GFMMNN \citep{gabrys2000general} as an example, the membership degree between a hyperbox and an input sample is computed based on the longest distance between that hyperbox and the input sample over all features. The smaller this distance is, the higher membership value is. Therefore, it is easily observed that the green hyperbox in Fig. \ref{expanation_2d} is closer to the input pattern than the blue hyperbox. As a result, the predicted class for the input pattern in this case is green.

\vspace{-0.45cm}
\begin{figure}[!ht]
    \centering
    \includegraphics[width=0.95\textwidth]{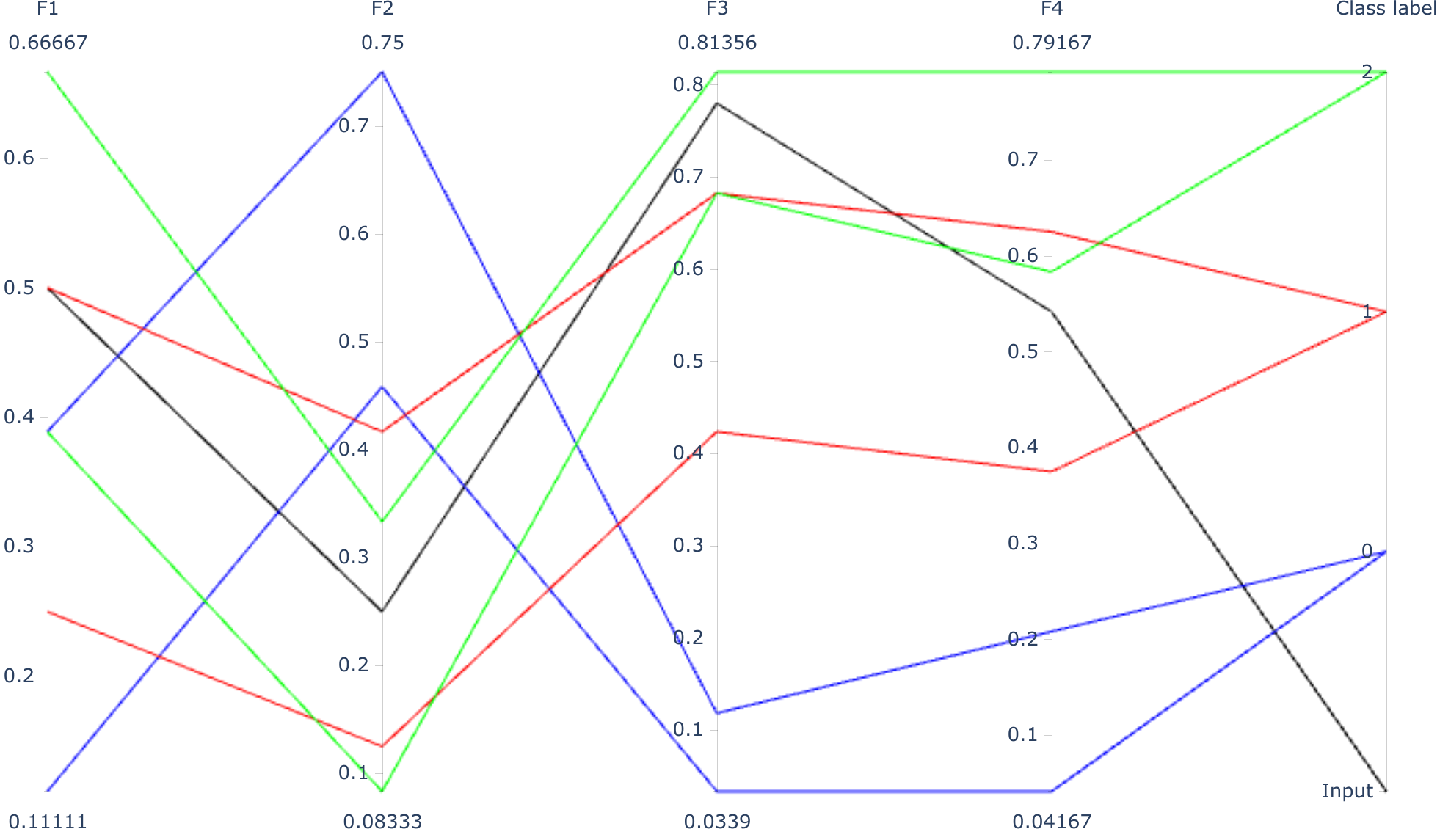}
    \caption{A parallel coordinates graph shows the representative (i.e. winning) hyperboxes for all classes in the context of an input pattern (black color) to be classified. In this case, the predicted class for the input pattern is \textit{green} based on the highest membership value.}
    \label{fig_3}
\end{figure}

The library also provides a general way of explanation for predicted outcomes using the parallel coordinates graph to visualise the coordinates of representative hyperboxes of all classes joining the prediction process as shown in Fig. \ref{fig_3}. In this case, the membership value between the green hyperbox and the input pattern is computed based on the fourth feature (F4), which shows the smallest distance compared to the distance values of other hyperboxes to the input pattern. Hence, the predicted class in this case is green.




\vskip 0.2in
\bibliography{main}

\end{document}